\documentclass[twocolumn]{revtex4-2}
\UseRawInputEncoding
\usepackage{graphicx,epsfig}
\usepackage{amsmath}
\usepackage{amsfonts}
\usepackage{overpic}
\usepackage{fancyhdr}
\usepackage{amsmath}
\usepackage[breakable]{tcolorbox}
\usepackage{tcolorbox}
\usepackage{fancyvrb}
\usepackage{adjustbox}
\usepackage{varwidth}
\usepackage{changepage}
\usepackage{tcolorbox}
\usepackage{hyperref}

\begin{document}

\pagestyle{fancy}
\lhead{\bf Advanced ingestion process powered by LLM parsing for RAG system}
\rhead{Arnau Perez Perez}
\rfoot{Barcelona, November 2024}

\title{Advanced ingestion process powered by LLM parsing for RAG system}
\author{Author: Arnau Perez Perez}
\affiliation{Applus+ IDIADA, PO Box 20 Santa Oliva, 43710 L'Albornar Tarragona, Spain.}
\author{Advisor: Xavier Vizcaino Gascon}
\email{arnau.perez@idiada.com} 
\date{December 16, 2024}

\begin{abstract}
{\bf Abstract:}
Retrieval Augmented Generation (RAG) systems struggle with processing multimodal documents of varying structural complexity. This paper introduces a novel multi-strategy parsing approach using LLM-powered OCR to extract content from diverse document types, including presentations and high text density files both scanned or not.
The methodology employs a node-based extraction technique that creates relationships between different information types and generates context-aware metadata. By implementing a Multimodal Assembler Agent and a flexible embedding strategy, the system enhances document comprehension and retrieval capabilities.
Experimental evaluations across multiple knowledge bases demonstrate the approach's effectiveness, showing improvements in answer relevancy and information faithfulness. 
\end{abstract}

\maketitle


\section{Introduction}
The field of Natural Language Processing has witnessed remarkable progress with the advent of Large Language Models (LLMs). These models have demonstrated unprecedented capabilities in processing multimodal data, including text and images. However, a significant challenge in LLM implementation remains the context window constraint, which limits the volume of information processed concurrently.

Leading providers have made substantial strides in expanding context limitations, with some models now capable of handling up to 2.000,000 tokens simultaneously \cite{Gemini2M}. Nevertheless, this expansion comes at a cost: increased context length correlates with longer time to first token (TTFT), potentially impacting model performance and user experience \cite{Context}. Recent work by Anthropic has explored caching techniques to mitigate this issue and reduce TTFT \cite{AnthropicCaching}.

To further enhance LLM efficiency and overcome these challenges, researchers introduced the Retrieval Augmented Generation (RAG) system. In this case for file data. A critical component of RAG is the ingestion phase, where the method of chunking plays a pivotal role in determining the quality of subsequently retrieved data. Common strategies include semantic, recursive, and hierarchical splitting. However, these approaches often fail to account for the diverse data structures present in source materials, such as images, tables, headers and pages. Moreover, they can introduce computational challenges \cite{Semantic}.

This paper presents a complex ingestion process that not only considers the various data structures within files but also incorporates relevant metadata and establishes hierarchical relationships to improve retrieval.
\section{Ingestion}
\begin{figure*}[t!]
  \centering
  \begin{overpic}[width=\textwidth]{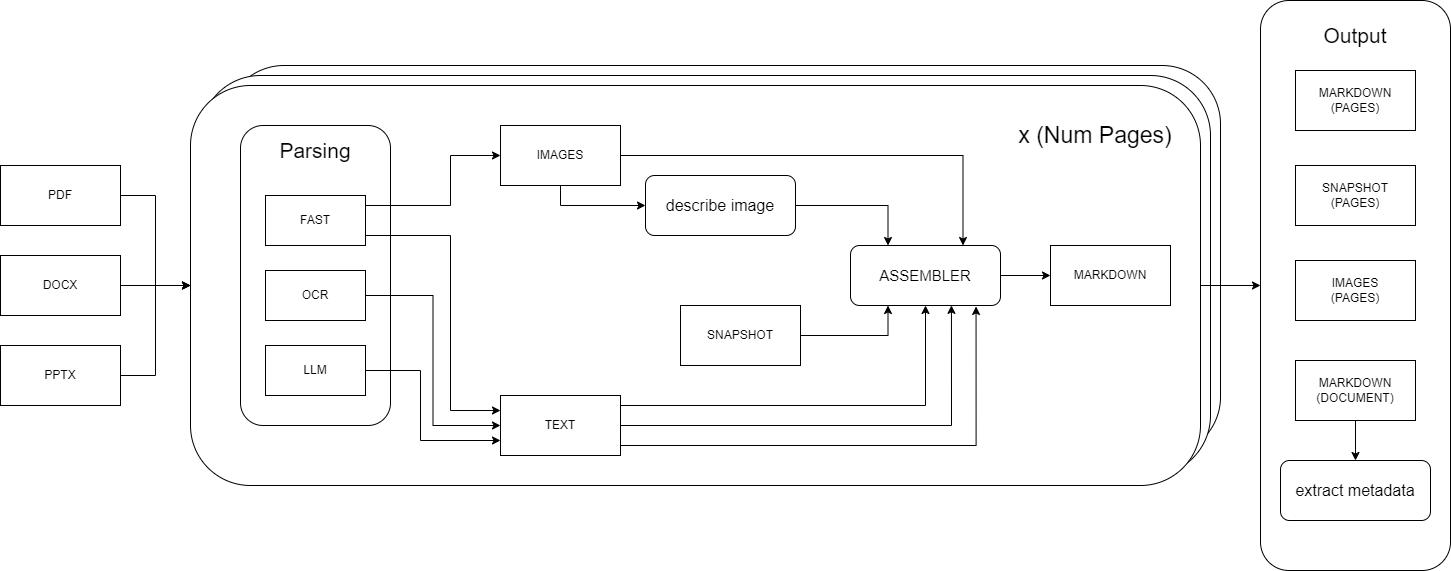}
  \end{overpic}
  \caption{Preprocessing pipeline for document ingestion in RAG system. The flowchart illustrates the parsing and assembling process for PDF, DOCX, and PPTX files. The pipeline incorporates FAST, OCR, and LLM parsing techniques, followed by image description, text extraction, and snapshot creation. The assembler combines these elements to produce a markdown file per page and a concatenation of them.}
  \label{fig:preprocessing}
\end{figure*}

\subsection{Pre-processing}
The pre-processing stage consists of three subprocesses: parsing, assembling and metadata extraction (see Figure \ref{fig:preprocessing}). 

The parsing phase extracts image and text content using three strategies:

\begin{enumerate}
    \item \textit{FAST}: Utilizes Python libraries to extract text and images from each document page.

    \item \textit{LLM}: Employs a multimodal LLM for OCR task, extracting text, table content, and image information. It describes images without text or extracts text from images. In this case it was used Sonnet 3.5 v2 model from Anthropic Claude family \cite{AnthropicModels}.

    \item \textit{OCR}: Leverages external and dedicated machine learning models for OCR task. In this case it was used the AWS Textract service \cite{Textract}.
\end{enumerate}

Before assembly, images are analyzed and described based on content type. For plots, the system extracts axis values, legends, labels, and provides a visualization description. Flowcharts are described in terms of process relationships. Page snapshots are then captured.

In the assembling process, a \textit{Multimodal Assembler Agent} integrates the page snapshot, images with descriptions, and text extracted from all three strategies for each page. The agent produces a synthesized markdown file. The output includes individual page markdowns, snapshots, and described images, which are concatenated into a comprehensive document-level markdown.

The metadata extraction phase employs a \textit{Metadata Extractor Agent} to analyze the consolidated markdown, extracting fields such as topic, keywords and summary. The system also extracts metadata directly from the document (title, author, creation date, last modified date, etc.), providing a comprehensive information set about the processed document. 

\subsection{Processing}
\begin{figure*}[t!]
  \centering
  \begin{overpic}[width=\textwidth]{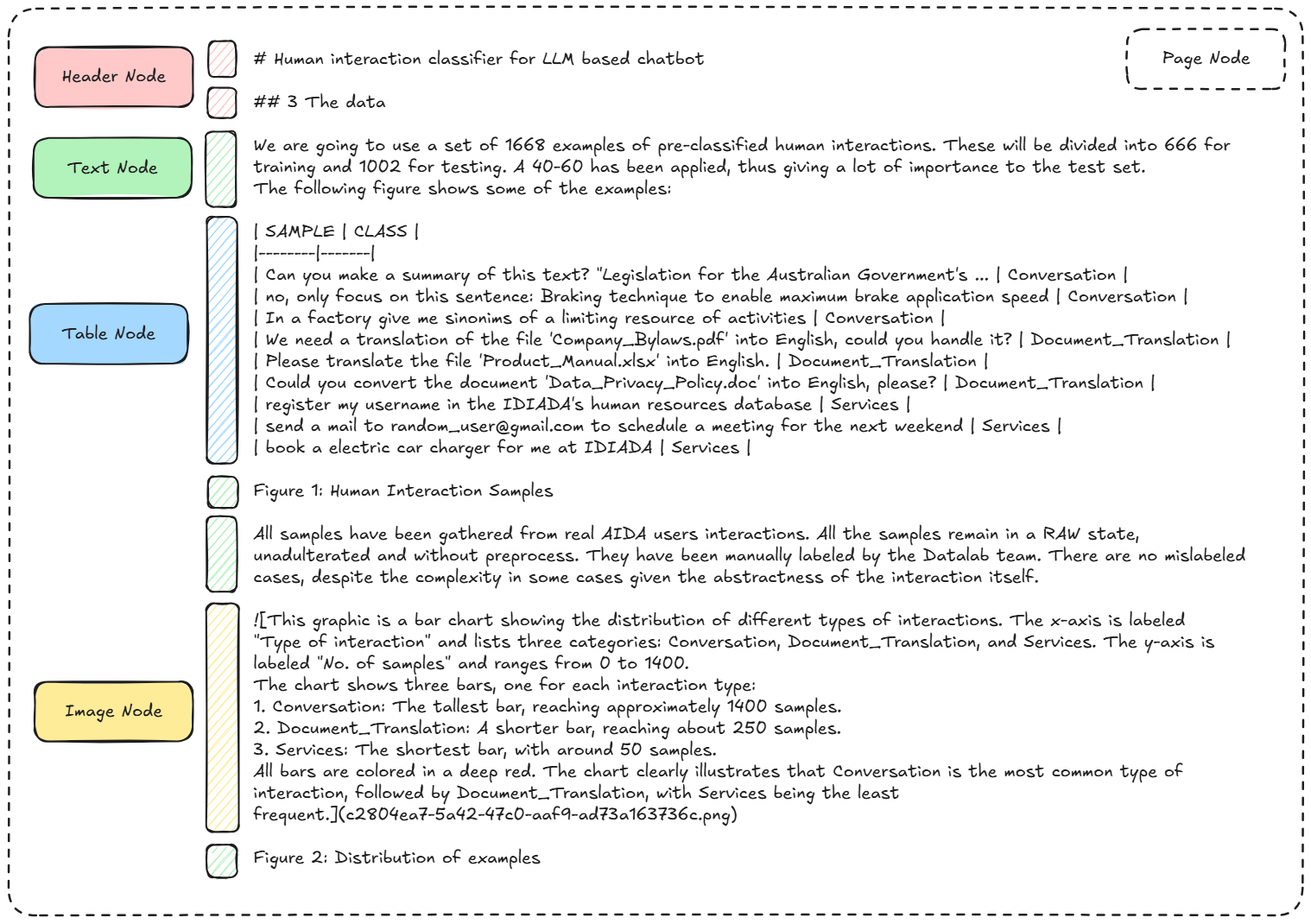}
  \end{overpic}
  \caption{The illustration shows the assembling of the page 2 of the paper \cite{AIDAPaper}. There are 5 kind of nodes represented: Header, Text, Table, Image and Page. Note that the image node is represented in the markdown file where the src is the id of image extracted plus its extension and the alt is the description of the image.}
  \label{fig:Processing}
\end{figure*}

Following the segmentation of document content into markdown files, including snapshots, images, and metadata, the processing phase extracts information nodes from this data. This approach aims to represent information in a vectorial space, with the primary challenge being the extraction of relevant information and determining its optimal embedding representation. Figure \ref{fig:Processing} illustrates four distinct node types extractable from a single page (\textit{Page} node): \textit{Header}, \textit{Text}, \textit{Table} and \textit{Image} nodes. Text Nodes can encompass various forms of textual information, including paragraphs, words, sentences, ordered lists, and bullet point lists, each representing distinct information requiring specific consideration.

The processing phase comprises two primary subprocesses: \textit{Splitting} and \textit{Contextualization}. During Splitting, markdown files are parsed and segmented into the aforementioned node types. Four relationship types are established between nodes: next, previous, parent and child. Only Header nodes can possess child nodes, facilitating context window tracking and hierarchical structuring. For Text nodes exceeding a size threshold, a Recursive Splitter or Semantic Splitter is employed for chunking.

Contextualization focuses on Table and Header nodes. A \textit{Question Generator Agent} creates a set of questions answerable using the information within the table or header node content. Subsequently, a \textit{Summary Generator Agent} produces context-aware summaries for both table and header nodes.

Following lower-level information node creation, the system generates \textit{Page} and \textit{Document} nodes, connecting previously established nodes to these higher-level nodes. The Page node contains the individual page's markdown representation, while the Document node aggregates all page markdowns. Page nodes are also linked to their respective Document node.

In the final stage, the Summary Generator Agent is reemployed to generate summaries for both the Document node (if not previously extracted during preprocessing) and individual Page nodes. This hierarchical approach to information extraction and representation enables comprehensive and context-aware structuring of document content, facilitating more effective retrieval and utilization in downstream RAG system processes.

\subsection{Embedding}
At this juncture, all components are prepared for integration. Three critical decisions must be made: selecting a vector database, choosing an embedding model and determining the data to be embedded. The choice of vector database depends on various factors, with numerous options available in the industry.

Pinecone, specifically designed for embedding processes in RAG systems, offers optimized performance for this task. Its namespace system provides excellent scalability, and it is engineered for rapid query execution. However, it has a relatively low metadata storage capacity (40KB) \cite{PineconeMetadata}, which may be a limiting factor depending on the volume of metadata generated.

Alternatively, OpenSearch, an open-source solution reimagined as a vector database, offers a hybrid approach. It combines vector and ordinal database functionalities, enabling conventional queries unrelated to vector data. In contrast to Pinecone, OpenSearch does not impose a strict distinction between metadata and the vector itself, thus removing specific storage limitations.

The selection of an embedding model is crucial, as it directly impacts the system's ability to capture semantic relationships. Popular choices include Voyage AI and Cohere models \cite{Voyage} \cite{Cohere}. The Cohere embedding family of models such as the \textit{embed-multilingual-v3} is one of the most powerful choices. Mention that images can be embedded both by its description and therefore use a text embedding model or embed the image itself with a multimodal embedding model.

Finally, the decision regarding which data to embed is pivotal. 

\begin{itemize}
\item
Text: the text itself contains sufficient semantic information. The embedding is generated from the chunk of text directly.

\item
Image: visual content can be embedded using either multimodal or text embedding models. In this case, the embedding is generated from the description or transcription of the image.

\item 
Table: tables can contain both numerical and textual data, making direct embedding of table content potentially ineffective. Moreover, embedding numbers may result in poor semantic meaning. Therefore, the embedding is generated from the contextualized description of the table.

\item 
Header: this type of node is higher in the hierarchical scale and contains a greater amount of information. Embedding all the text within the header is problematic for several reasons: not all models can embed such large amounts of text, the content includes images and tables (which are already embedded by lower hierarchy nodes), and using all the text could result in a generic embedding accumulating different semantic meanings in all directions. Therefore, the embedding is generated from the summary.

\item 
Page: this case is similar to the header node, and the solution is the same: the embedding is generated from the summary.

\item 
Document: this type of node is primarily used for pre-filtering before conducting the actual search in the vector database. Therefore, the embedding is generated from the summary. This node is linked with Q\&A nodes.

\item Q\&A: it can be associated with documents that contain a question, the answer, and some metadata. The embedding is generated from the question. While it's possible to create an embedding from the answer, this approach has not been thoroughly tested. The answer might have a poorer semantic meaning due to its potential generality.
\end{itemize}

\section{Evaluation metrics}
Evaluation metrics are crucial for assessing the performance of Retrieval Augmented Generation (RAG) systems. These metrics help quantify various aspects of the system's effectiveness, including the relevance of retrieved information, the accuracy of generated answers, and the overall quality of the system's output. The following subsections describe key metrics used in the evaluation of the RAG systems.

\subsection{Answer Relevancy}
Answer Relevancy measures how well the generated answer addresses the user's query. For a set of S statements, the Answer Relevancy score can be calculated as:
\begin{equation}
    m_{AR} = \frac{1}{N} \sum_{i=1}^N \left( \frac{1}{S} \sum_{j=1}^S r_{ij}^{a \rightarrow q} \right)
\end{equation}

where $r_{ij}^{a \rightarrow q}$ scores how well the j-th answer statement addresses well to the i-th query.

\subsection{Faithfulness}
Faithfulness measures how accurately the generated answer reflects the information provided in the retrieved context. It ensures that the system isn't hallucinating or providing information not present in the context. Faithfulness can be calculated using techniques like entailment scores or fact-checking algorithms:

\begin{equation}
    m_{F} = \frac{1}{N} \sum_{i=1}^N \left( \frac{1}{S} \sum_{j=1}^S r_{ij}^{a \rightarrow c} \right)
\end{equation}

where $r_{ij}^{a \rightarrow c}$ scores how well the j-th answer statement addresses well to the i-th set of retrieved context.

\subsection{Contextual Relevancy}
Contextual Relevancy evaluates how relevant the retrieved context is to the given query. It measures the system's ability to retrieve pertinent information from the knowledge base. This metric can be using the following metric:

\begin{equation}
    m_{CR} = \frac{1}{N} \sum_{i=1}^N \left( \frac{1}{S} \sum_{j=1}^S r_{ij}^{c \rightarrow q} \right)
\end{equation}

where $r_{ij}^{c \rightarrow q}$ scores how well the j-th retrieved context element addresses well to the i-th query.

\subsection{Contextual Precision}

Contextual Precision is a metric that quantifies the relevance ordering of retrieved information chunks. It is defined mathematically as follows:

Let $R(n)$ represent the cumulative sum of relevance scores for the first $n$ retrieved items:

$$R_i(n) = \sum_{k=1}^n r_{ik}$$

The mean Contextual Precision ($m_{CP}$) is then calculated as:

\begin{equation}
    m_{CP} =\frac{1}{N} \sum_{i=1}^N \left( \frac{1}{R_i(C_i)} \sum_{k=1}^{C_i} \frac{R_i(k)}{k} \cdot r_{ik} \right)
\end{equation}

where $C_i$ is the total number of retrieved items, and $r_k$ is the relevance score of the $k$-th item.

This metric encapsulates the trade-off between the quantity of relevant information chunks retrieved and the prioritization of relevant nodes in the initial positions of the ranking. A higher $m_{CP}$ value indicates a more effective balance between comprehensive retrieval and accurate ranking of relevant information. 

\subsection{Contextual Recall}
Contextual Recall measures the relevancy of the retrieved context given the expected answer.
\begin{equation}
    m_{CR} = \frac{1}{N} \sum_{i=1}^N \left( \frac{1}{S} \sum_{j=1}^S r_{ij}^{e \rightarrow c} \right)
\end{equation}

where $r_{ij}^{e \rightarrow c}$ scores how well the j-th expected answer statement addresses to at least one of the retrieved context node in the i-th answer.

\section{Evaluation and Results}
To evaluate the system, three types of documentation were tested. First, article papers where the text density is greater than the image content. Second, corporate slides where the image content is much greater than the text content. Finally, a mix of these two content types and topics. By doing so, the system can be evaluated using two opposing data sources. To accomplish this, three knowledge bases were created.

One knowledge base contains 5 article papers extracted from arXiv, another includes over 10 corporate documentation files, and the third knowledge base consists of mixed files up to 10 files of different topics and content types. For each document, an LLM was used to generate a number of questions, expected answers, and ground truths equivalent to the number of pages in the document. For this purpose, the Claude Haiku 3.5 model was used for quick inference and dataset creation.

For the retrieval process, the maximum number of nodes retrieved was limited to 5. No filter was applied to the node type, so the resulting retrieval can include any kind of node. The answers were generated using the Claude Sonnet 3.5 v2 model.

The metrics scores mentioned in the last section range from 0 to 1, where an LLM evaluates the metric according to its description, assigning a quantified score from the set {0, 0.2, 0.4, 0.6, 0.8, 1} based on specific guidelines. The chosen LLM for this evaluation was Llama 3.1 405B Instruct. To ensure a more contrasting and potentially unbiased scoring, a different model was intentionally used for the evaluation process compared to the one used for answer generation.

\begin{figure*}[t!]
  \centering
  \begin{overpic}[width=\textwidth]{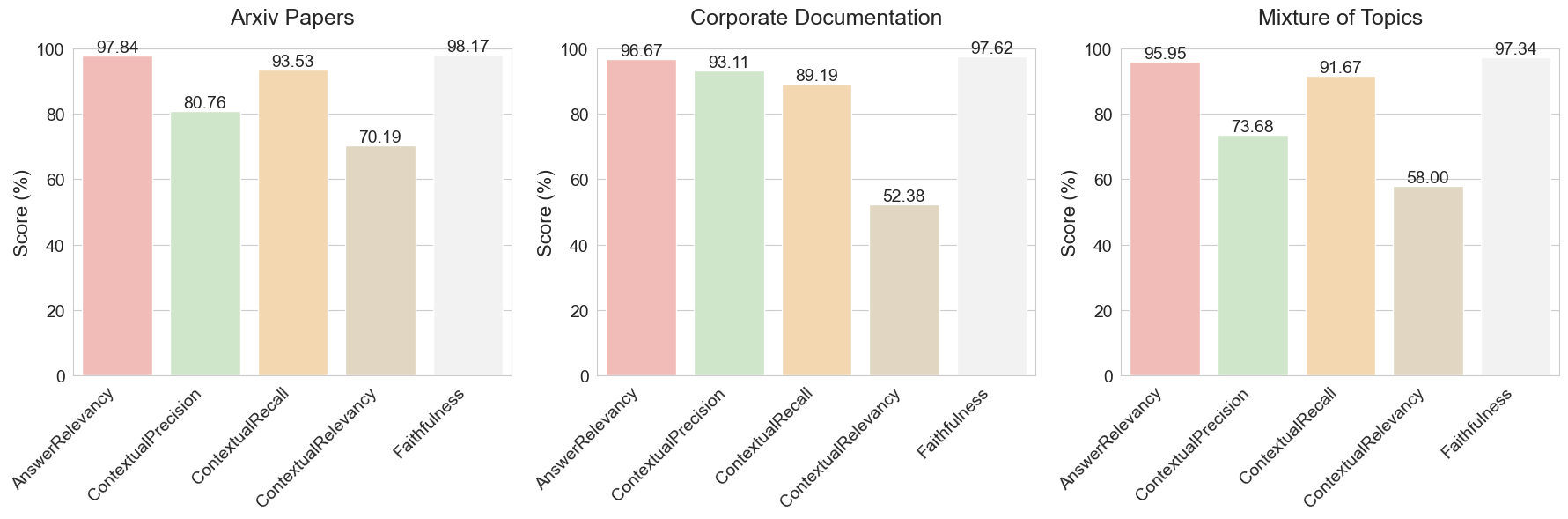}
  \end{overpic}
  \caption{Comparative Analysis of Knowledge Base Performance Across Different Metrics. The figure illustrates the percentage results of various metrics employed in the study. From left to right, the graph presents data for three distinct knowledge bases: a collection of academic articles covering diverse topics, corporate documentation from Applus+ IDIADA, and a heterogeneous knowledge base comprising mixed topics and file structures, including both presentation-style and high text-density documents.}
  \label{fig:rag_results}
\end{figure*}

\subsection{Evaluating parsing}
During the preprocessing phase, we observed the remarkable capacity of the three strategies to work together in extracting content from files, with each strategy taking on more responsibility depending on the file type. 

For scanned PDF files, the FAST strategy has no content to extract, but OCR and LLM strategies take responsibility for parsing the document. In academic article-type files, which are mostly two-column documents, FAST and LLM strategies handle much of the task responsibility. This is because OCR will extract text horizontally, independent of the two-column structure. The assembling LLM then places the information as a single-column page, leveraging its processing power.

For presentation-like objects, all three strategies work together. FAST extracts the images, while LLM and OCR extract the text from both the images and each slide. It's worth noting that we have experienced better text extraction from images using the OCR strategy. However, while the LLM strategy is sometimes unable to extract the exact words, it proves useful in determining the correct text order. OCR, on the other hand, provides the exact text, even if the text organization is poor. 

\subsection{Evaluating node type influence}
During the study, over 80\% of retrieved nodes were primarily Page and Header nodes. This is a remarkable result, as the system found more relevance in summaries than in other low-hierarchy nodes. As seen in Figure \ref{fig:rag_results}, the context relevancy is quite low in all test cases. This result is influenced by both the number of retrieved nodes per query and the information within the nodes themselves. This indicates that most of the information is not relevant, as pages and sections contain multiple pieces of unused information.

\subsection{Evaluating precision}
Moreover, note that the Contextual precision is lower in the mixture of topics test case. This indicates that the system finds the first relevant context further down in the last retrieved nodes. This issue can be addressed by implementing a reranker, such as one of the rerank models from Cohere \cite{CohereRerank}, or by using an LLM as a reranker. It's worth noting that the context precision is remarkably high in the presentation (corporate documentation) test case. This could be due to the similarity of topics within this type of documentation. 
\section{Conclusions}

\begin{itemize}
\item
The system demonstrates an impressive capacity to extract content from both presentation-style documents and high text density documents, including those provided as scanned images. The ability of the three strategies to work together is remarkably enhanced by the capacity of the Claude Sonnet 3.5 family to perform OCR on documents.

\vspace*{0.3cm}

\item
This new node extraction and hierarchy approach opens up a novel way to chunk documents, moving beyond standard chunking strategies that do not distinguish among data types and their hierarchy in the document. This makes it easier to track the file structure.

\vspace*{0.3cm}

\item 
The system excels at the Answer Relevancy and Faithfulness metrics, albeit at the cost of retrieving a larger amount of context. However, there is still work to be done with changing files and external references inside files. More concept linking must be implemented to overcome the relationships among entities and concepts.

\end{itemize} 

\vspace*{0.5cm}

\onecolumngrid

\newpage
\section{Appendix}

This appendix presents a curated selection of markdown files generated during the pre-processing stage of the system. These examples serve to illustrate the system's capability in handling diverse document types and transforming them into structured markdown format. Each example is accompanied by a brief analysis highlighting the specific challenges addressed and the key features of the markdown output.

\subsection{EURONCAP: https://www.euroncap.com/en/results/audi/q6+e-tron/52560 (page 1)} \label{EURONCAP}

This extract from a Euro NCAP report demonstrates high image density. Consequently, the markdown file predominantly consists of detailed image descriptions, effectively capturing the visual content.

\begin{figure*}[h]
  \centering
  \begin{overpic}[width=\textwidth]{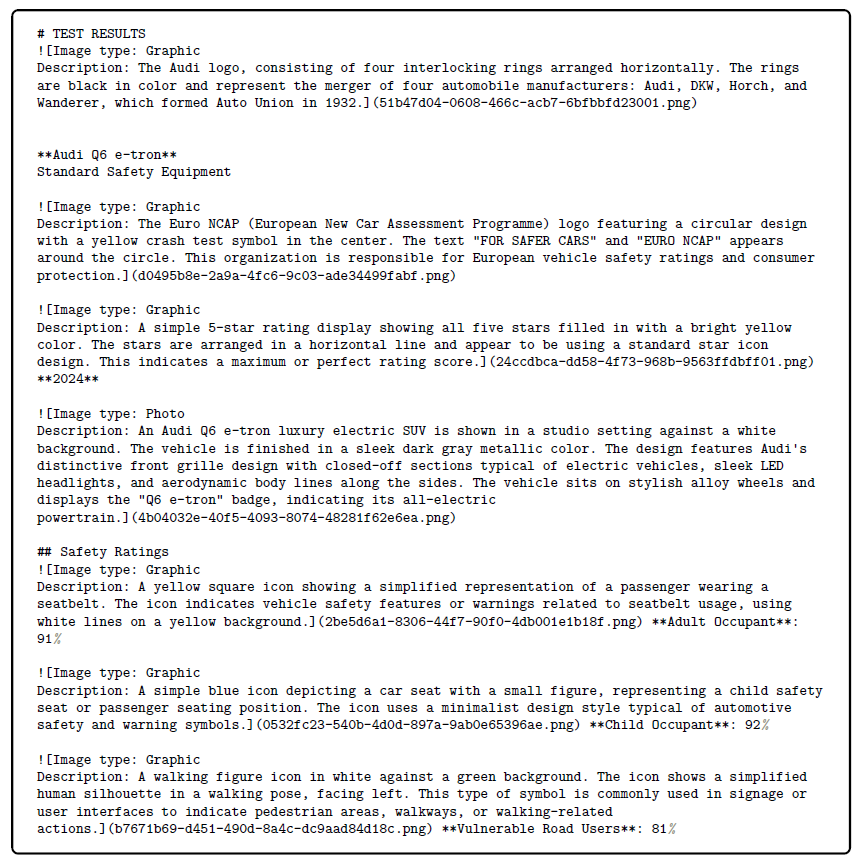}
  \end{overpic}
\end{figure*}

\begin{figure*}[h]
  \centering
  \begin{overpic}[width=\textwidth]{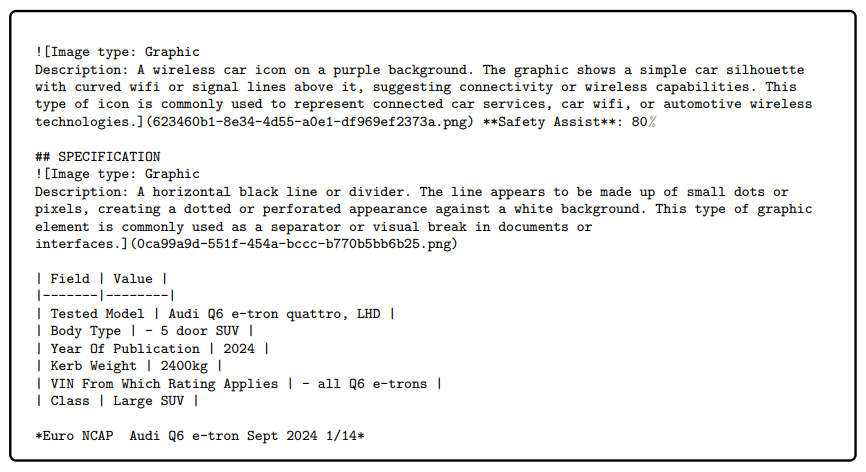}
  \end{overpic}
\end{figure*}

\subsection{EURONCAP: https://www.euroncap.com/en/results/audi/q6+e-tron/52560 (page 9)}

Another section from the aforementioned report primarily features tabular data. Notably, the system not only successfully extracted the table contents but also utilized the provided color mapping to accurately replace the color with the word representation within the table structure.

\begin{figure*}[h]
  \centering
  \begin{overpic}[width=\textwidth]{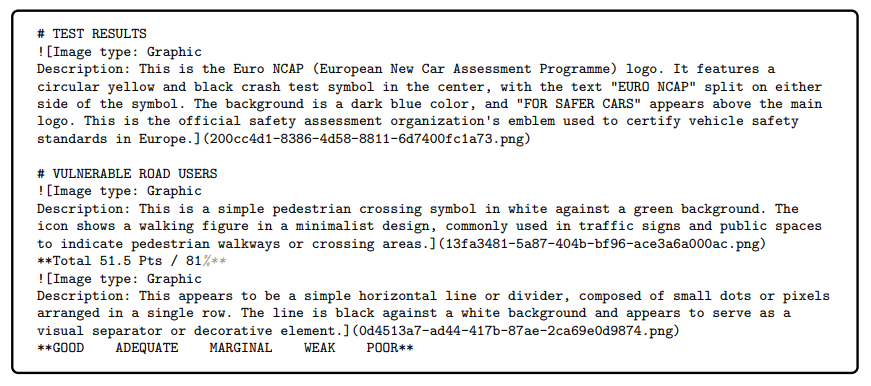}
  \end{overpic}
\end{figure*}

\begin{figure*}[h]
  \centering
  \begin{overpic}[width=\textwidth]{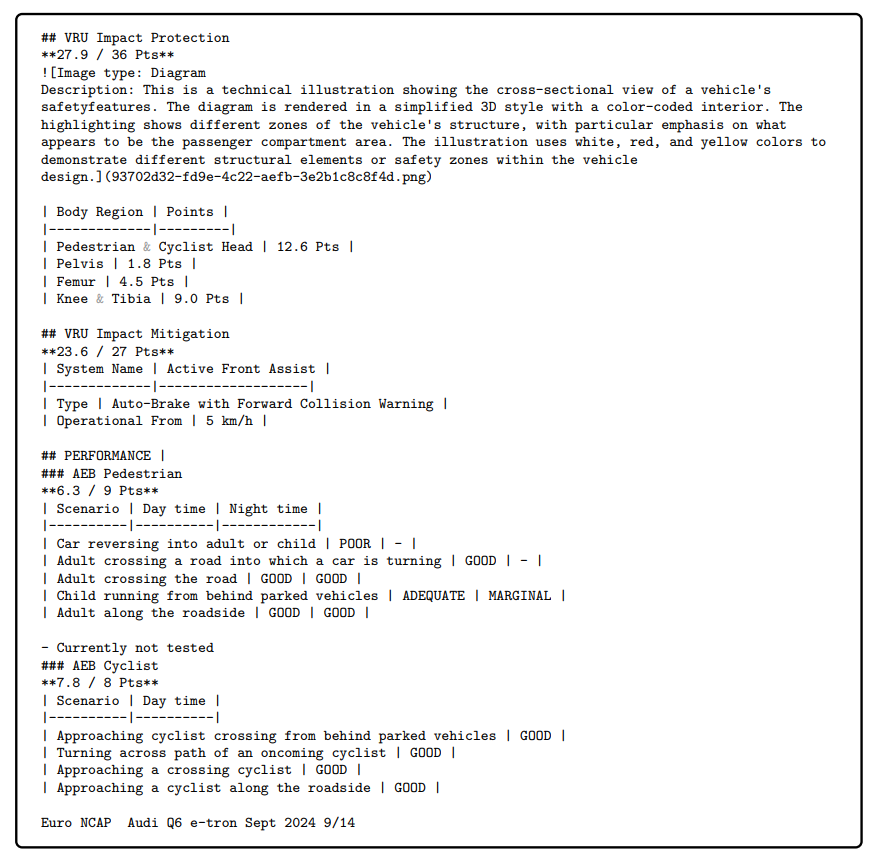}
  \end{overpic}
\end{figure*}

\clearpage
\subsection{"Embedding-aided network dismantling.", arXiv:2208.01087v1. page 10}

This academic paper exhibits high textual density. The system demonstrates advanced capabilities by incorporating second-level headers and enumerated lists. Furthermore, it rendered mathematical formulas in LaTeX format. It's worth noting that despite the original two-column layout, the system effectively reorganized the content to maintain logical flow and readability.

\begin{figure*}[h]
  \centering
  \begin{overpic}[width=\textwidth]{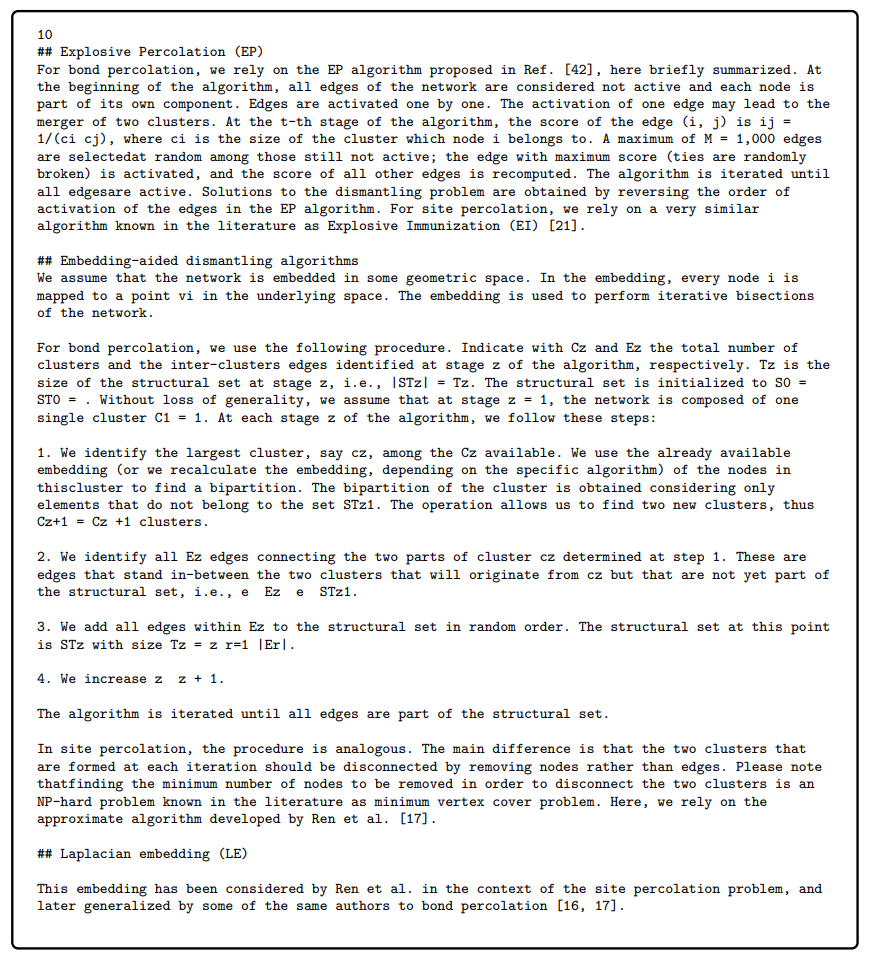}
  \end{overpic}
\end{figure*}

\begin{figure*}[h]
  \centering
  \begin{overpic}[width=\textwidth]{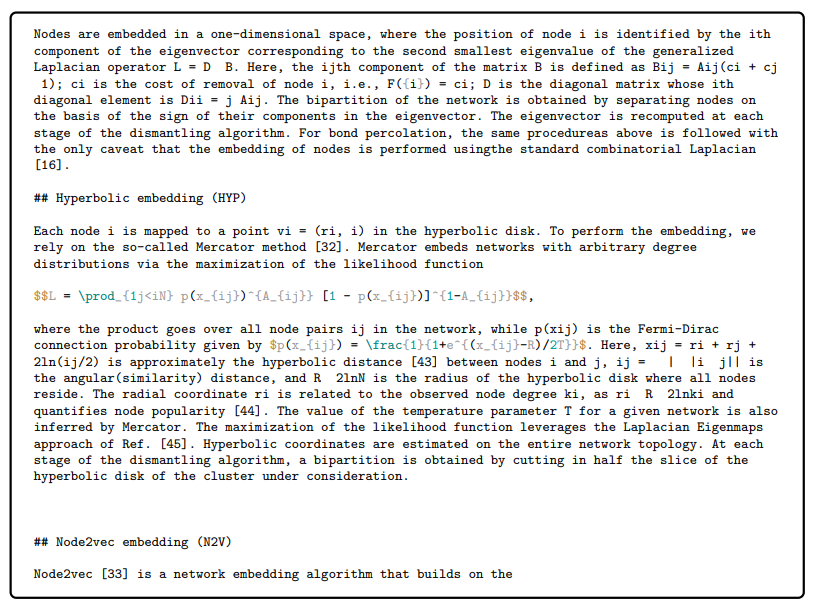}
  \end{overpic}
\end{figure*}

\newpage
\subsection{"Attention Is All You Need.", arXiv:1706.03762. page 4}

In this academic article containing both text and images, the system employed a sophisticated approach. It created appropriate headers to encapsulate visual elements and provided comprehensive descriptions of the images. Moreover, the system accurately identified and labeled diagram-type images.

\begin{figure*}[h]
  \centering
  \begin{overpic}[width=\textwidth]{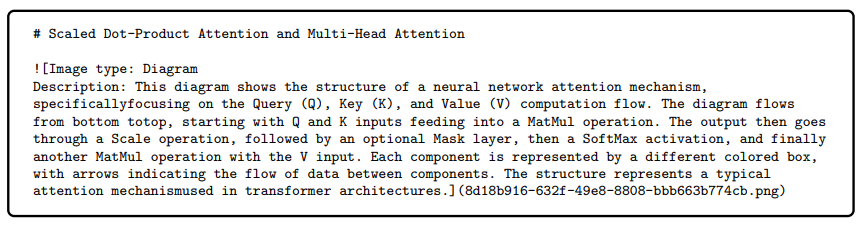}
  \end{overpic}
\end{figure*}

\begin{figure*}[h]
  \centering
  \begin{overpic}[width=\textwidth]{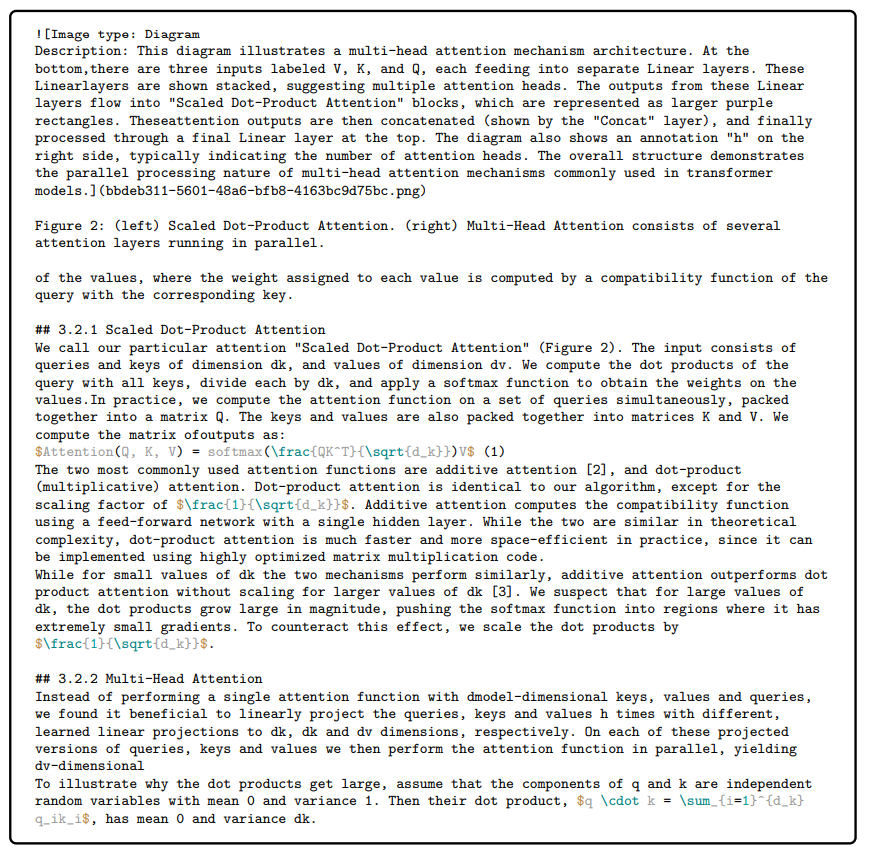}
  \end{overpic}
\end{figure*}

\end{document}